\documentclass{article}
\usepackage{graphicx} 
\usepackage{acl}
\usepackage{url}
\usepackage{multirow}
\usepackage{enumitem}
\usepackage[final]{microtype}

\title{Connecting Ideas in `\textit{Lower-Resource}' Scenarios: NLP for National Varieties, Creoles and Other Low-resource Scenarios}
\author{Aditya Joshi$^1$ Diptesh Kanojia$^2$ Heather Lent$^3$ Hour Kaing$^4$ Haiyue Song$^4$ \\
  $^1$University of New South Wales, Australia \\
  $^2$Institute for People-Centred AI, University of Surrey, United Kingdom\\
  $^3$Department of Computer Science, Aalborg University, Denmark\\
  $^4$National Institute of Information and Communications Technology, Japan \\
  \texttt{aditya.joshi@unsw.edu.au}, \texttt{d.kanojia@surrey.ac.uk},
  \texttt{hcle@cs.aau.dk}\\
  \texttt{\{hour\_kaing,haiyue.song\}@nict.go.jp}\\
}
\date{May 2024}

\begin{document}

\maketitle
\begin{abstract}
\textit{(Selected as a tutorial at COLING 2025)} Despite excellent results on benchmarks over a small subset of languages, large language models struggle to process text from languages situated in `lower-resource' scenarios such as dialects/sociolects (national or social varieties of a language), Creoles (languages arising from linguistic contact between multiple languages) and other low-resource languages. This \textbf{introductory tutorial} will identify common challenges, approaches, and themes in natural language processing (NLP) research for confronting and overcoming the obstacles inherent to data poor contexts.  By connecting past ideas to the present field, this tutorial aims to ignite collaboration and cross-pollination between researchers working in these scenarios. 
Our notion of `lower-resource' broadly denotes the outstanding lack of data required for model training - and may be applied to scenarios apart from the three covered in the tutorial.
\end{abstract}

\section{Introduction}
Contemporary NLP systems benefit from enormous data available for a small minority of the world's languages. However, a lack of resources persists for many languages or even forms of languages. The people most affected by such data sparsity speak widely-spoken languages from socioeconomically disadvantaged localities, or languages that develop as a combination of multiple languages. These languages collectively represent `lower-resource' scenarios. Our choice of the term `lower' covers more than the traditionally understood notion of low-resource languages. We identify three lower-resource scenarios: national varieties (including dialects or sociolects), Creoles, and other low-resource languages. The challenges of engineering effective NLP systems for such \textit{lower-resource} scenarios require careful planning at every stage of NLP research, to overcome the problem of data sparsity. 

While each of the lower-resource scenarios bears its unique socio-historical contexts, the tutorial brings together researchers working separately in these scenarios. Collectively, the tutorial will connect past research in terms of: 
\begin{itemize}[noitemsep,topsep=1pt]
  \item Challenges in data curation  
  \item Potential for wide linguistic variation (\textit{e.g.,} existing on a linguistic continuum or eschewing strict spelling conventions, \textit{etc.})
  \item Need for smart modeling choices over greedy ones 
  \item Increased model vulnerability
\end{itemize}

This introductory tutorial identifies the emergence of `lower-resource' scenarios, specifically national varieties, Creoles and other low-resource languages, and highlights commonalities and differences in the research across the three areas. We expect the tutorial to enable attendees to identify these connections or learn from insights obtained in a related scenario. In addition to stimulating ideas for new approaches in the respective scenarios, the tutorial will enable a better understanding of the transfer learning approach and community mobilization towards equitable language technologies that help maintain global linguistic diversity.

\section{Target Audience \& Impact}
The tutorial will be developed for an audience interested in familiarizing themselves with past research in NLP for lower-resource scenarios. They would be expected to be broadly familiar with typical NLP tasks, models, and techniques. The tutorial will be designed for two key personas: (a) graduate students working in one of the scenarios; and (b) Industrial practitioners who wish to adapt their current NLP tools to markets where dialects, Creoles, or other lower-resource languages are used. Senior researchers and academics may also be able to identify insightful connections in research methodologies of the scenarios covered in the tutorial. The tutorial will foster a deeper understanding and collaboration among researchers working on lower-resource scenarios. By highlighting common challenges and solutions across different tasks, the tutorial will enable participants to leverage insights from related research areas, thereby enhancing the development of more effective and inclusive NLP technologies, while contributing to the promotion of linguistic diversity.   

\section{Outline}
We propose a \textbf{full-day} in-person tutorial at COLING 2025. The tutorial outline below is mapped to the following learning objectives (LO). At the end of the tutorial, participants will be able to:
\begin{itemize}[label={}]
\item \textbf{LO1:} Comprehend techniques across natural language understanding and generation  for dialects and low-resource languages.
\item \textbf{LO2:} Experiment with popular techniques using datasets and sample code provided.
\item \textbf{LO3:} Apply techniques for their research.
\item \textbf{LO4:} Appreciate the relationship between the scenarios from a computational perspective.
\end{itemize}
The tutorial is divided into six modules, the tentative outline of which is as follows:
\begin{itemize}[leftmargin=*, label={}] 
    \item \textbf{Module 1: Introduction} (Expected time: 30 minutes; Teaching Artifacts: Slides; LO4)\\
    This module discusses recent developments in NLP, and establishes the motivation and structure for this tutorial.
     \begin{itemize}
        \item Transformer \& Language Models
        \item Dialects, Creoles, and other lower-resourced languages 
        \item Motivation \& Computational tasks
        \item Objectives \& Structure of the tutorial.
    \end{itemize}
     \item \textbf{Module 2: Emerging Connections} (Expected time: 60 minutes; Teaching Artifacts: Slides, Hands-on; LO1, LO2 and LO4)\\
     This module introduces the scenarios, relevant language families, and common challenges.
        \begin{itemize}
            \item Linguistic Considerations (Variability and informality of dialects; Phylogenetic relationships of low-resource languages, etc.)
            \item Resource issues (Scarcity, data quality, annotation)
            \item Fine-tuning \& Prompting as key access points
            \item Emerging common themes in the tutorial
         \item Hands-on Session (10 mins)
        \begin{itemize}
        \item Evaluate zero-shot on a dialect dataset for sarcasm detection highlighting results, challenges, and pitfalls.
        \end{itemize}
        \item Q\&A \& Discussion (10 mins)
        \end{itemize}

    \item Coffee Break: 30 minutes
    \item \textbf{Module 3: Common Ideas in Dataset Creation} (Expected time: 40 minutes; Teaching Artifacts: Slides, Hands-on; LO2 and LO3)\\
    This module discusses data curation strategies with example case studies.
    \begin{itemize}
        \item Identifying data sources
        \item Computationally assisted annotation tools.
        \item Leveraging multilingual datasets, code-mixing
        \item Case studies (Synthetic Data, Translation, and Resource Creation)
        \item Hands-on Session (10 mins)
        \begin{enumerate}
        \item Data annotation exercise with the audience
        \item Discuss results and challenges/pitfalls
        \end{enumerate}
        \item Q\&A \& Discussion (10 mins)
    \end{itemize}
    \item \textbf{Module 4: Common Themes in Understanding} (Expected time: 90 minutes; Teaching Artifacts: Slides, Hands-on; LO1 and LO2)\\
    This module discusses language understanding tasks and challenges specific to linguistically diverse lower-resource scenarios.
    \begin{itemize}        
        \item Language Understanding basics
        \item Sequence Classification (60 mins)
            \item Task-specific Case Studies
            \begin{enumerate}
                \item Sentiment analysis for three scenarios
                \item Aggression detection for three scenarios
            \end{enumerate}
        \item Hands-on Session (30 mins)
        \begin{enumerate}
            \item Implementing text classification model for LRL
            \item Discussion on model performance.
        \end{enumerate}

        \item Token Classification (30 mins)
        \begin{enumerate}
            \item Overview of techniques for sequence classification tasks with POS tagging and related tasks (such as NER) as case studies
            \item Comparison of approaches and performance metrics
        \end{enumerate}
        \item Q\&A \& Discussion(10 mins)        
    \end{itemize}
    \item Lunch Break, duration as decided by conference organizers
    \item \textbf{Module 5: Common themes in Generation} (Expected Time: 90 mins; Teaching Artifacts: Slides, Hands-on; LO1, LO2))\\
    This module discusses language generation tasks and their challenges in lower-resource scenarios with some existing solutions.
    \begin{itemize}
    \item Tasks, Challenges \& Solutions (40 mins)
        \begin{enumerate}
            \item Tasks: machine translation, text summarization, and dialogue generation.
            \item Challenges: 1) data scarcity, 2) low subword coverage, and 3) evaluation metrics.
            \item Solutions: 1) data augmentation including back-translation and subword regularization, and transfer learning from related languages, 2) script normalization, and 3) robust metrics such as xCOMET and chrF.
        \end{enumerate}


   \item Generation tasks for dialects (20 mins)
       \begin{enumerate}
            \item Tasks \& challenges
            \item Techniques including dialect normalization, data creation, adaptation, and dialect-aware generation
       \end{enumerate}
    
    \item Hands-on Session (20 mins)
        \begin{enumerate}
        \item Training machine translation systems for English$\rightarrow$Hindi using the Airavata framework\footnote{\url{https://ai4bharat.github.io/airavata}}
        \item Evaluating and refining generated content
        \end{enumerate}
    \item Q\&A \& Discussion (10 mins)
    \end{itemize}
    
    \item Coffee Break: 30 minutes


     \item \textbf{Module 6: Conclusion and Q\&A} (Expected time: 50 minutes; Teaching Artifacts: Slides, Hands-on; LO1, LO4)
     \begin{enumerate}
        \item Video snippets from researchers who work on one or more of the language groups in focus. 
        \item Recap of key concepts
        \item Discussion with the audience using prompt questions.
        \item Future work in terms of potential collaborations between researchers working on low-resource scenarios.
     \end{enumerate}
\end{itemize}
All linguistic examples used during the presentation and the hands-on sessions will either be in English or will be accompanied by 
English translations.
\section{Diversity Considerations}
The topic of the tutorial highlights the effort to diversify the scope of NLP research beyond the predominant languages. This includes but is not limited to dialects of major languages such as English and Japanese, and low-resource languages such as Hindi and Marathi. The topic is particularly relevant for researchers who speak or work on datasets in these languages, where techniques that rely on large volumes of data may not be applicable. 

The tutorial brings together the experiences of the presenters working separately on NLP for lower-resource scenarios. The presenters work in diverse geographies, speak several languages and language varieties, and are all early to mid-career researchers. The presenters have worked in different areas of NLP such as sentiment analysis, parsing, and machine translation.
\section{Reading List}
\begin{itemize}
\item Joshi, A., Dabre, R., Kanojia, D., Li, Z., Zhan, H., Haffari, G., \& Dippold, D. (2024). Natural language processing for dialects of a language: A survey. arXiv preprint arXiv:2401.05632.
\item Kanojia, D., Dabre, R., Dewangan, S., Bhattacharyya, P., Haffari, G. and Kulkarni, M., 2020, December. Harnessing Cross-lingual Features to Improve Cognate Detection for Low-resource Languages. In Proceedings of the 28th International Conference on Computational Linguistics (pp. 1384-1395).
\item Lent, H., Tatariya, K., Dabre, R., Chen, Y., Fekete, M., Ploeger, E., ... \& Bjerva, J. (2023). CreoleVal: Multilingual Multitask Benchmarks for Creoles. arXiv preprint arXiv:2310.19567. (Accepted at TACL)
\item Gala, J., Jayakumar, T., Husain, J.A., Khan, M.S.U.R., Kanojia, D., Puduppully, R., Khapra, M.M., Dabre, R., Murthy, R. and Kunchukuttan, A., 2024. Airavata: Introducing hindi instruction-tuned llm. arXiv preprint arXiv:2401.15006.
\item Deoghare, S., Choudhary, P., Kanojia, D., Ranasinghe, T., Bhattacharyya, P. and Orašan, C., 2023, July. A Multi-task Learning Framework for Quality Estimation. In Findings of the Association for Computational Linguistics: ACL 2023 (pp. 9191-9205).
\end{itemize}
\section{Presenters}
\textbf{Dr. Aditya Joshi} (he/him) is a Lecturer/Assistant Professor in the School of Computer Science \& Engineering at the University of New South Wales, Sydney, Australia, and specializes in natural language processing (NLP). His current research interests lie in NLP for dialects of English, and optimization of NLP models. Aditya has presented two pre-conference tutorials at *ACL conferences (EMNLP 2017: Computational Sarcasm with Prof. Pushpak Bhattacharyya, IIT Bombay; and AACL 2020: NLP for Healthcare in the Absence of a Healthcare Dataset with Sarvnaz Karimi, CSIRO Data61, Australia). Aditya teaches fundamental (data structures and algorithms; typical class sizes: 500) as well as specialized (natural language processing, artificial intelligence; typical class sizes: 100) courses at UNSW Sydney, a large, research-intensive university. He was acknowledged as an outstanding reviewer at ICML, ACL, EACL and IJCAI. His publications have been in ACL, EMNLP, COLING, WWW, and so on, and have received 2600+ citations and best paper awards at ACM FAccT, ACM MoMM and so on. Google Scholar: \url{https://scholar.google.com/citations?hl=en&user=SbYRrvgAAAAJ}

\textbf{Dr. Diptesh Kanojia} (he/him) is a Lecturer/Assistant Professor at Institute for People-centred AI, School of Computer Science and Electrical Engineering, at the University of Surrey, United Kingdom. He has been working on various Natural Language Processing sub-domains and his current research interests are quality estimation (QE), automatic post-editing (APE), social NLP and low-resource NLP scenarios. Previously, Diptesh has presented a tutorial on 'Unsupervised Neural Machine Translation' at ICON 2020. He regularly publishes and reviews at the *ACL, AAAI, IJCAI, and ECCV conferences, and has been awarded best paper mention at EACL 2021 for his work titled ``Cognition-aware Cognate Detection''. He is also a co-organiser for the Conference for Machine Translation (WMT) Shared tasks on QE and APE. He leads the taught module on NLP at the University of Surrey. Additionally, he is passionate about teaching, diversity \& inclusion, and applying NLP research to real-world applications. Google Scholar: \url{https://scholar.google.com/citations?user=UNCgCAEAAAAJ}

\textbf{Dr. Heather Lent} is a postdoctoral researcher at Aalborg University in Copenhagen, Denmark. She received her PhD in 2022 from the University of Copenhagen in Computer Science, where she first began her work in Creole NLP. Prior to this, she has several years experience working in bio/medical NLP, another under-resourced domain. Inasmuch, her primary research interests pertain to low-resource scenarios within NLP. Specifically, she is interested in working to strengthen areas of NLP which could be described as \textit{vulnerable}, whether that be NLP for vulnerable languages (e.g. Creoles), or addressing the security vulnerabilities inherent to the NLP models (e.g., embedding inversion attacks), so that reliable and safe NLP can be available to all.  She has publications in TACL, ACL, EMNLP, COLING, LREC, and *CL workshops. 
Google Scholar: \url{https://scholar.google.com/citations?user=ZxdpWo4AAAAJ}

\textbf{Dr. Hour Kaing} is a researcher at the Advanced Translation Technology Laboratory, National Institute of Information and Communications Technology (NICT), Japan. He received his M.Sc from University of Grenoble 1, France, and his Ph.D. from NARA Institute of Science and Technology, Japan. He is interested in linguistic analysis, machine translation, language modeling, and speech processing. He has previously presented a tutorial on ``Linguistically Motivated Neural Machine Translation'' at EAMT 2024. He has publications in TALLIP, EACL, PACLIC, LREC, and IWSLT, and has reviewed papers for TALLIP, LREC-COLING, and ICON. Google Scholar: \url{https://scholar.google.com/citations?user=zJ0wGWoAAAAJ&hl}

\textbf{Dr. Haiyue Song} (haiyue.song@nict.go.jp) is a technical researcher at the Advanced Translation Technology Laboratory, National Institute of Information and Communications Technology (NICT), Japan. He obtained his Ph.D. at Kyoto University. He has presented \href{https://github.com/prajdabre/eamt24-linguistic-mt}{a tutorial at EAMT 2024}. His research interests include machine translation, large language models, subword segmentation, and decoding algorithms. He has MT and LLMs-related publications in TALLIP, AACL, LREC, ACL, and EMNLP. Additional information is available at \url{https://shyyhs.github.io}.

\section{Other Information}
We expect around 20\% of the conference attendees who would consider attending a tutorial to be interested in ours. This could potentially be more considering the accessibility of the location of COLING 2025. The tutorial will require a room with a projector for the presenters, and wireless internet for all participants.
\section{Ethics Statement}
The tutorial will cover past research highlighting key papers, creating summary tables and insights for the audience. Every module will be followed by a timeslot for questions and discussion. The last module will have dedicated time for audience discussion using a set of initial prompt questions. These discussions will ensure that alternative perspectives are shared in order to augment the material covered in the tutorial. The tutorial will include illustrative examples in languages other than English.
\end{document}